\documentclass{article} 
\usepackage{iclr2027_conference,times}
\renewcommand{\iclrruler}[1]{}

\usepackage{amsmath,amsfonts,bm}

\def\eqref#1{equation~\ref{#1}}

\def\1{\bm{1}}

\DeclareMathAlphabet{\mathsfit}{\encodingdefault}{\sfdefault}{m}{sl}
\SetMathAlphabet{\mathsfit}{bold}{\encodingdefault}{\sfdefault}{bx}{n}

\usepackage{hyperref}
\usepackage{url}
\usepackage{booktabs}
\usepackage{amsmath}
\usepackage{latexsym}

\usepackage{graphicx}
\usepackage{subcaption}

\usepackage{booktabs}
\usepackage{multirow}

\usepackage[T1]{fontenc}
\usepackage{microtype}
\usepackage{inconsolata}

\usepackage{xurl}
\usepackage{float}
\usepackage{enumitem}
\usepackage{float}
\usepackage{tcolorbox}
\tcbuselibrary{raster}
\usepackage{enumitem}
\usepackage{wrapfig}
\usepackage{amsmath}
\usepackage{booktabs}
\usepackage{tabularx}
\usepackage{tcolorbox}
\tcbuselibrary{breakable,skins}
\usepackage{hyperref}

\newtcolorbox{memorycase}{
    colback=white,
    colframe=black!60,
    boxrule=0.4pt,
    arc=0pt,
    outer arc=0pt,
    boxsep=0pt,
    left=5pt,
    right=5pt,
    top=5pt,
    bottom=5pt,
    fontupper=\footnotesize,
    valign=top,
    before upper={
        \setlength{\parindent}{0pt}
        \setlength{\parskip}{3pt}
    }
}

\newlist{caseitems}{itemize}{1}
\setlist[caseitems]{
    label=\textbullet,
    leftmargin=1.2em,
    itemsep=3pt,
    topsep=2pt,
    parsep=0pt,
    partopsep=0pt
}

\title{OA-MAP: Evidence-Grounded Multi-Agent Multimodal Framework for Interpretable Knee Osteoarthritis Progression}

\author{
\parbox{\textwidth}{\centering
Sixu Chen$^{1,2}$ \quad
Mingrui Yang$^{1}$ \quad
Qiang Guan$^{2}$ \quad
Xiaojuan Li$^{1}$
\\[0.5em]
{\normalfont
$^{1}$ Cleveland Clinic, Cleveland, OH, USA \\
$^{2}$ Kent State University, Kent, OH, USA
}
}
}

\date{}

\iclrfinalcopy 
\begin{document}

\maketitle
\fancyhead{}
\renewcommand{\headrulewidth}{0pt}

\begin{abstract}
Knee osteoarthritis (KOA) progression prediction can support patient monitoring, requiring the integration of multimodal data and multidomain expertise. Moreover, isolated risk estimates provide limited insight underlying a prediction. To automate the progression assessment workflow and reduce manual effort while providing interpretable findings and supporting evidence, we present OA-MAP, an autonomous multi-agent framework for evidence-grounded assessment of structural and pain progression in KOA. The system incorporates modality-specific agents including MRI, X-ray, and clinical agents, together with a coordinator agent. This framework can autonomously recruit specialist agents, select tools for prediction and analysis, and retrieve literature as external evidence based on user request and available patient information. An uncertainty-informed human-in-the-loop mechanism enables clinicians to review and correct intermediate findings, triggering recomputation of affected results. We evaluate the prediction models using 600 participants from the FNIH Osteoarthritis Biomarkers Consortium cohort. On the test set of 100 participants, the fusion models achieve AUROCs of 0.80 for structural progression and 0.68 for pain progression. A case study illustrates how OA-MAP combines risk estimates with intermediate findings, cross-modal conflicts, literature support, and uncertainty indicators to support interactive review.
\end{abstract}

\section{Introduction}



Knee osteoarthritis (KOA) is a prevalent degenerative joint disease, affecting an estimated 364.6 million people worldwide in 2019~\cite{yang2023burden,wang2025survey}. Early identification of patients at risk of progression can inform patient monitoring and support participant selection for clinical trials~\cite{tiulpin2019multimodal}. Previous studies have investigated the prediction of structural and pain progression using imaging and clinical data, including multimodal approaches~\cite{tiulpin2019multimodal,panfilov2025end}. Despite the effictiveness of these methods, they primarily focus on improving final prediction accuracy, while overlooking the intermediate decision-making process, clinical consistency, and interpretability of the results. Without this information, clinicians may find it difficult to assess the basis and reliability of an individual result. Practical implementation presents another challenge. Using these models often requires preparing model-specific inputs, selecting an appropriate model for the available modalities and intended prediction endpoint, and coordinating multiple processing steps. These tasks require technical expertise and additional effort, creating barriers to routine clinical use. Together, these challenges motivate an integrated framework that simplifies model execution while providing interpretable predictions, traceable evidence, and explicit uncertainty assessment.



Large language model (LLM) agents offer a means of coordinating such workflows by combining language-based reasoning with tool use. These agents can interpret user requests, select and execute actions, and incorporate external information into subsequent decisions~\cite{yao2022react}. Applications include collaborative biological research~\cite{swanson2025virtual}, tool-assisted chest radiograph interpretation~\cite{fallahpour2025medrax}, and the selection and execution of clinical risk calculators~\cite{jin2025agentmd}. These developments suggest the potential of agent-based interfaces to connect clinical requests with specialized computational tools. Multi-agent systems extend this approach by distributing tasks among agents with distinct responsibilities, such as data analysis, evidence retrieval, and result verification. This division of labor can facilitate the integration of heterogeneous data and specialized tools, while allowing agents to review intermediate outputs and identify inconsistencies across different stages of a workflow.

However, reliability remains a central challenge. LLM agents can generate unsupported statements, misinterpret clinical information, and use tools incorrectly. Evaluations in clinical decision-making settings have documented errors in data interpretation and instruction following, including attempts to invoke nonexistent tools~\cite{hager2024evaluation}. Multi-agent collaboration does not inherently overcome these limitations. Consensus-seeking discussion can induce conformity, causing agents to abandon initially correct judgments and converge on a shared error~\cite{cui2026free}. Moreover, when agents are differentiated primarily through role prompts, their apparent diversity may not translate into independent reasoning or complementary evidence, particularly if they share the same underlying model and information. Agreement among agents may therefore reflect shared biases or propagated errors and should not, by itself, be treated as independent corroboration. These limitations motivate agent architectures that combine clearly defined responsibilities with verifiable evidence, explicit handling of disagreement, and transparent uncertainty assessment.

To address these challenges, we propose a multi-agent framework for evidence-grounded assessment of structural and pain progression in KOA. The framework integrates specialized prediction models, automated workflow orchestration, literature-based evidence assessment, and clinician review. Given a patient ID and a clinical request, our framework retrieves patient records from a memory module and coordinates the relevant modality-specific agents. Each specialist selects and executes tools appropriate to the available data and requested target endpoint. This automation reduces the need for users to manually prepare inputs and coordinate model execution.

Each specialist grounds its assessment in modality-specific model outputs and retrieves relevant PubMed literature to examine whether published associations between patient features and progression support or conflict with the relationships learned by the models. Specialists are differentiated through their data modalities, computational tools, and associated modality-specific literature evidence, providing a substantive basis for complementary assessments. The coordinator obtains the final progression label from a fusion model using all available modality data and reports it alongside modality-specific predictions, model explanations, literature assessments, and uncertainty indicators.

By presenting intermediate findings, model explanations, and traceable literature evidence alongside KOA progression predictions, the framework enables clinicians to examine the basis of each assessment while helping to limit unsupported agent interpretations.

We further introduce an uncertainty score that combines model-derived uncertainty indicators, disagreement across modalities, and literature-based evidence assessments to comprehensively evaluate the reliablity of the prediction result. This score also guides a human-in-the-loop module in which clinicians can inspect intermediate findings and correct suspected errors. The agents then propagate these revisions through the affected predictions, literature assessments, uncertainty calculations, and final report. This interactive process allows clinicians to examine the basis of a prediction and assess how corrections to intermediate findings change the final assessment.

Our contributions are fourfold:
\begin{itemize}
    \item We provide interpretable assessments of structural and pain progression in KOA by integrating model-based explanations with external literature evidence. The resulting reports present intermediate findings, supporting evidence, conflicts, and evidence gaps, offering clinicians a more comprehensive view of each patient's predicted progression risk and the basis for the assessment.

    \item We develop an agent-based workflow that automates patient-data retrieval, modality- and endpoint-specific tool selection, prediction, and report generation in response to clinical requests, reducing the need for manual coordination across separate processing pipelines.

    \item We design modality-grounded coordination that differentiates agents through distinct source data inputs, specialized tools, and modality-specific evidence pools. This grounds agent specialization in their information sources and capabilities, rather than role descriptions alone, with the aim of preserving modality-specific evidence and complementary assessments.

    \item We incorporate uncertainty-informed human-in-the-loop that enables correction of intermediate findings when uncertainty score is high. These revisions trigger recomputation of affected predictions, literature assessments, uncertainty indicators, and reports, allowing clinician feedback to directly influence the final assessment.
    
\end{itemize}


\section{Related Work}

\subsection{LLM Agent}

LLM-based agents have been investigated as coordinators of specialist reasoning and external tools in biomedical research and healthcare~\cite{wu2023autogen}. In biological discovery, the Virtual Lab organizes an LLM principal investigator and specialist agents into a research team, with human feedback guiding the development of a computational pipeline for nanobody design \cite{swanson2025virtual}. In medical reasoning, MedAgents uses role-based expert discussions to develop and revise answers to medical questions \cite{tang2024medagents}. MDAgents further adapts the collaboration structure to task complexity, selecting between individual and group reasoning \cite{kim2024mdagents}.

Another line of work connects language agents to executable medical tools. MMedAgent learns to select specialized tools for tasks across multiple medical modalities \cite{li2024mmedagent}. MedRAX integrates chest X-ray analysis tools with multimodal LLMs and dynamically invokes these tools to address medical queries \cite{fallahpour2025medrax}. AgentMD curates executable clinical calculators from PubMed and selects and applies relevant calculators to patient information for risk assessment \cite{jin2025agentmd}~\cite{xiong2024benchmarking}. These systems~\cite{joseph2025machine,leung2020prediction,joseph2022machine} provide a foundation for coordinating prediction models, image analysis, and external knowledge within an agent workflow. Building on these capabilities, our framework focuses on KOA progression assessment, where agents coordinate modality-specific analysis and assess the literature support for influential feature--progression relationships.

\subsection{Knee osteoarthritis }
Knee osteoarthritis (KOA) prediction research has showned that multi-modal data provide complementary information for predicting structural and symptomatic progression~\cite{hu2023deepkoa,li2024predictive}. Integrating these modalities has therefore become an important direction in KOA progression modeling~\cite{tiulpin2019multimodal,panfilov2025end,cai2020comparison}.

Interpretability has also received explicit attention. Logistic-regression models with nomograms have been developed for separate radiographic and pain progression outcomes \cite{li2024predictive}. DeepKOA uses Grad-CAM to visualize image regions associated with its predictions \cite{hu2023deepkoa}. Explainable machine learning has further combined quantitative MRI and clinical variables with SHAP, regression coefficients, and permutation importance to examine influential predictors \cite{harari2025predicting}. These methods expose model associations and salient findings. Our framework extends this model-level interpretability by evaluating feature--progression relationships against external literature and examining agreement and conflicts across modality-specific predictions, providing a more comprehensive assessment of the evidence underlying each prediction.

\section{Method}

\begin{figure}[htbp]
    \centering
    \includegraphics[width=0.95\linewidth]{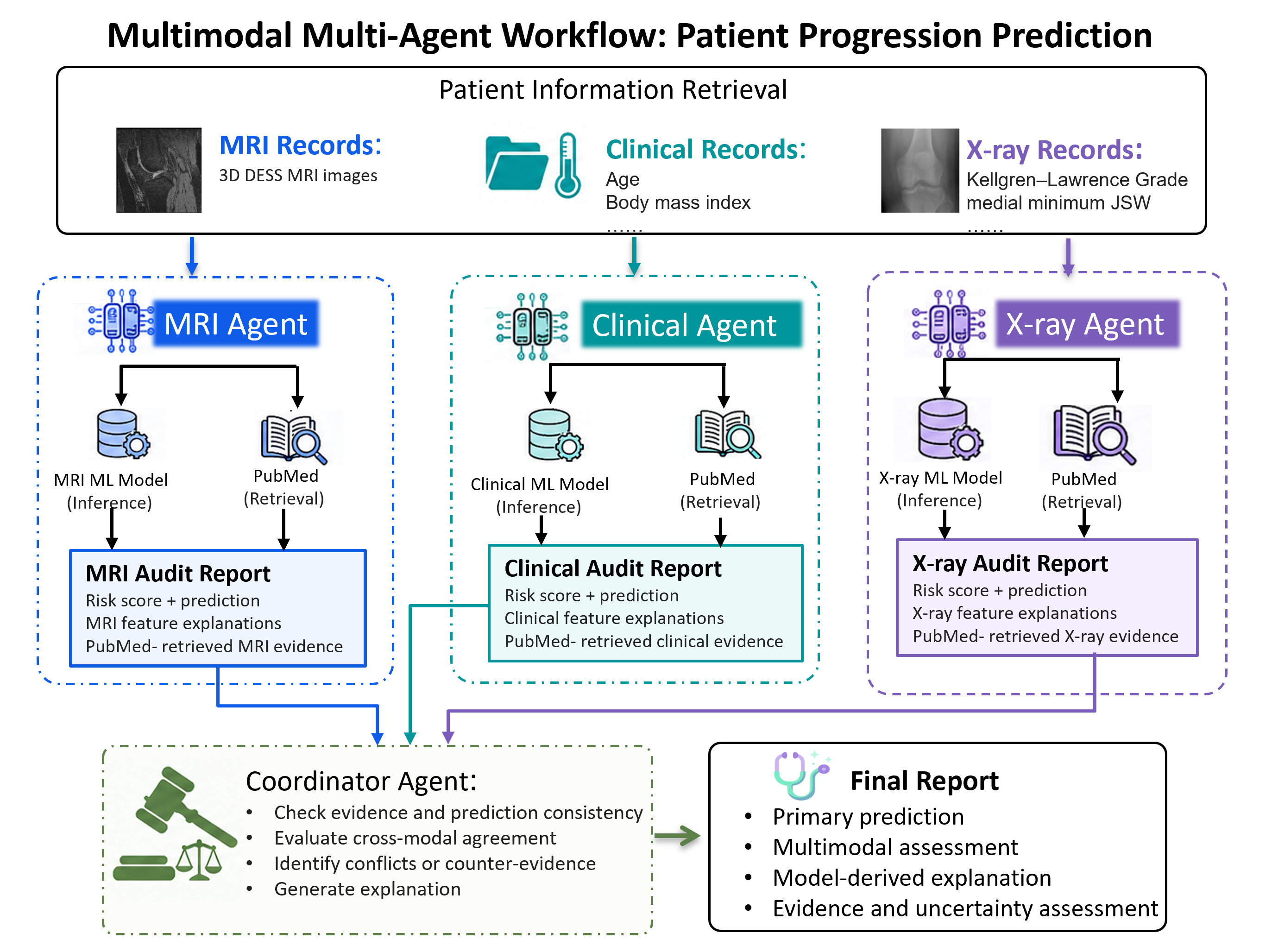}
    \caption{Overview of the proposed multi-agent framework
    for knee osteoarthritis progression assessment.}
    \label{fig:workflow}
\end{figure}

Figure~\ref{fig:workflow} illustrates the overall architecture of the proposed framework. Given a patient ID and a clinical request, our framework automatically retrieves available patient records from the
memory module. Based on the requested task and available modalities, the framework activates the appropriate modality-specific agents and supplies each with the required inputs. Each specialist agent performs
modality-specific analysis, prediction, and literature
evidence assessment. The processed modality agent outputs are then passed to the coordinator for fusion-model inference, uncertainty assessment, and final structured report generation.


\subsection{Memory Module}
The memory module organizes patient information into structured records indexed by patient ID. Clinical records include age, sex, body mass index, symptom and functional assessments, and medication-use information. Imaging records contain 3D DESS MRI images, together with X-ray-derived radiographic measurements. Each record retains knee laterality, visit information, and modality availability, enabling our framework to retrieve the relevant data for the requested prediction task.

\subsection{Modality-specific Agent}

Our framework includes three modality-specific agents including clinical agent, MRI agent and X-ray agent. Each agent receives the corresponding modal patient data retrieved from the memory module. It invokes a logistic-regression model appropriate to its own modality and the requested target, then retrieves external literature to assess the relationships underlying the prediction. For a given endpoint, let \(\mathbf{z}_m\) denote the preprocessed feature vector for modality \(m\), and let \(f_m\) denote the selected prediction model:
\[
p_m=f_m(\mathbf{z}_m),
\qquad
\hat y_m=\mathbb{I}[p_m\geq\tau_m],
\]
where \(p_m\) is the model's progression score, \(\tau_m\) is its training-derived decision threshold, and \(\mathbb{I}[\cdot]\) is the indicator function. The label \(\hat y_m=1\) denotes a predicted progressor for the selected endpoint, whereas \(\hat y_m=0\) denotes a predicted non-progressor.

Literature retrieval focuses on features with the largest contributions to the patient's prediction. For the logistic-regression models, feature importance is measured by \(|\beta_{mj}z_{mj}|\), where \(\beta_{mj}\) is the coefficient of feature \(j\). This ranking guides top-\(K\) feature selection under the predefined retrieval budget. The selected features are mapped to clinical concepts, anatomical terms, and endpoint-specific PubMed queries. Queries expand from specific feature--anatomy combinations to broader predictor concepts, and article titles and abstracts are retrieved through NCBI E-utilities.

Findings are assessed for comparability in measurement, anatomy, temporal role, and prediction endpoint. Those failing these criteria are labeled \emph{not comparable}, while comparable findings with mixed, statistically nonsignificant, or directionally unclear associations are labeled \emph{inconclusive}. For the remaining findings, association directions are aligned with the model's feature and outcome coding and compared with the corresponding logistic-regression coefficient signs. Concordant and opposing directions are labeled \emph{supporting} and \emph{conflicting}, respectively.

These assessments characterize literature support for the model's feature relationships and inform the coordinator's evidence-based uncertainty calculation. By linking model-derived feature associations to verifiable literature evidence, this process provides clinicians with a traceable basis for interpreting patient-specific progression risk and identifying discrepancies that warrant further clinical review.

\textbf{MRI agent.}
The MRI agent automatically coordinates the tools required to convert a 3D DESS MRI into an endpoint-specific progression assessment. It manages the dependencies between image analysis, feature extraction, prediction, and literature retrieval, using intermediate tool outputs to determine the appropriate inputs for subsequent stages. Given an MRI volume, the agent first invokes segmentation tools to identify 17 anatomical ROIs: six femoral regions, two patellar regions, seven tibial regions, and the medial and lateral menisci.

The agent uses each eligible ROI's anatomical label to automatically select and invoke the appropriate abnormality classifier. It routes the fourteen cartilage ROIs to the cartilage classifier and the two meniscal ROIs to the meniscal-morphology classifier. For each selected crop, the agent uses the Qwen2.5-VL-3B-Instruct visual encoder and supplies the resulting representation, together with the region identity, to the appropriate classifier. For crop \(I_r\) and abnormality field \(k\),
\[
\mathbf{h}_r=V(I_r),
\qquad
q_{rk}=
\operatorname{softmax}\!\left(C_k(\mathbf{h}_r,r)\right)_1,
\]
where \(V\) denotes the visual encoder and \(C_k\) produces binary presence logits. The cartilage classifier returns two scores per ROI, representing nonzero cartilage-size and cartilage-depth grades. The meniscal classifier returns three morphology scores per ROI, corresponding to the anterior horn, body, and posterior horn. The agent retains these 34 continuous abnormality-presence scores for progression prediction and uses validation-derived thresholds to identify regions for visual review.

To complement the classifier outputs, the MRI agent invokes quantitative measurement tools on the segmentation results. These tools produce four directional meniscal-extrusion measurements and 16 geometric descriptors, comprising cartilage volume and thickness descriptors across four compartments, two joint-gap measurements, and two meniscal-coverage ratios. The agent assembles their outputs into a common representation:
\[
\mathbf{x}_{\mathrm{MRI}}
=
[\mathbf{q}_{34};\mathbf{e}_{4};\mathbf{g}_{16}]
\in\mathbb{R}^{54}.
\]

Conditioned on the requested progression target, the agent selects the corresponding prediction model and supplies \(\mathbf{x}_{\mathrm{MRI}}\) in the required feature order. The agent then uses the model's feature contributions to guide literature retrieval and assess external support for the influential feature--progression relationships. It forwards the prediction, MRI findings, and literature assessments to the coordinator for integration.

The MRI agent supports human-in-the-loop refinement by incorporating clinician corrections to ROI-level abnormality scores. Following each correction, it updates the corresponding input features and reruns the affected prediction and literature-assessment steps. The revised outputs are then propagated to uncertainty assessment and final report generation.

\textbf{X-ray Agent.}
The X-ray agent receives radiographic information, comprising 18 baseline features that describe global and compartmental OA grades, joint-space measurements, osteophytes, sclerosis, and alignment. Based on the requested structural or pain target, the agent automatically selects the corresponding prediction model and supplies the features in the required order.

The agent uses the prediction model's feature contributions to guide subsequent evidence retrieval in PubMed literature. It identifies influential radiographic features, maps them to clinical concepts, and invokes the PubMed retrieval tool with queries matched to the anatomical compartment and prediction target. The retrieved findings are interpreted and compared with the model's feature relationships to identify supporting, conflicting, or inconclusive evidence. The agent then forwards a structured modal output containing the prediction score, decision threshold, predicted label, feature explanations, and assessed literature evidence to the coordinator agent.

\textbf{Clinical Agent.}
The clinical agent takes clinical demographic data as input. Its input comprises six baseline variables: age, sex, body mass index, target-knee WOMAC pain, WOMAC physical-function limitation, and frequent knee-pain medication use. Given the requested target, the agent automatically selects the corresponding clinical prediction model and invokes the further PubMed-based evidence retrieval.

\subsection{Coordinator Agent}
The coordinator agent invokes an endpoint-specific fusion model using all available modalities and integrates structured outputs from the available modality-specific agents to produce the final assessment. 

It then performs three important tasks. First, it identifies disagreements among modality-specific predictions and conflicts between model relationships and retrieved PubMed evidence and quantifies uncertainty from these complementary sources. Second, it flags uncertain or conflicting findings that warrant clinician review, providing the relevant predictions and evidence for human assessment. Third, it consolidates the integrated prediction, modality-specific findings, literature assessments, and uncertainty estimates into a structured final report.

\textbf{Uncertainty Quantification.}
The coordinator agent quantifies three sources of uncertainty: proximity to the prediction boundary, disagreement between modality-specific predictions, and insufficient or conflicting literature evidence. Let \(\mathcal{M}\) denote the selected modalities, and let \(p_m\), \(\tau_m\), and \(\hat y_m\) denote the prediction score, classification threshold, and predicted label returned by modality-specific agent \(m\), respectively.

Prediction uncertainty is determined by the distance between the model score and its classification threshold. We normalize this distance by \(1-\tau_m\) for scores at or above the threshold and by \(\tau_m\) for scores below it. Uncertainty is then defined as one minus this normalized distance:

\begin{equation}
\begin{aligned}
&u_m =
\begin{cases}
1-\dfrac{p_m-\tau_m}{1-\tau_m},
& p_m\geq\tau_m,\\[6pt]
1-\dfrac{\tau_m-p_m}{\tau_m},
& p_m<\tau_m,
\end{cases}
\\[10pt]
&U_{\mathrm{pred}} =
\frac{1}{|\mathcal{M}|}
\sum_{m\in\mathcal{M}}u_m.
\end{aligned}
\end{equation}

The uncertainty score \(u_m\) is highest (\(u_m=1\)) when the model score equals the classification threshold. As the score moves away from the threshold toward 0 or 1, uncertainty decreases and model confidence increases. The coordinator averages these uncertainty scores across all selected modalities to obtain \(U_{\mathrm{pred}}\), where a higher value indicates greater average uncertainty in the modality-specific predictions.

Modality-conflict uncertainty measures disagreement among all available modality-specific agents. The coordinator compares each pair of agents and averages their conflict scores:
\begin{equation}
U_{\mathrm{conflict}}
=
\frac{1}{|\mathcal{P}|}
\sum_{(m,n)\in\mathcal{P}}
\mathbb{I}\!\left[\hat y_m\neq\hat y_n\right]
\left|p_m-p_n\right|,
\end{equation}
where \(\mathcal{P}\) contains all distinct pairs of agents with available predictions. A pair contributes zero when its predicted labels agree. When the labels disagree, its contribution is the absolute difference between the two prediction scores. Thus, the overall conflict score is zero when all agents predict the same label, while higher values indicate stronger disagreement among their predictions. This component is calculated whenever at least two agents provide predictions; otherwise, it is recorded as unavailable.

Literature-evidence uncertainty summarizes the support for the feature relationships underlying each modality-specific prediction. Let \(\mathcal{K}_m\) contain the \(K\) features with the largest absolute log-odds contributions, \(a_{mj}=\beta_{mj}z_{mj}\). For feature \(j\), let \(N_{mj}\) denote the number of eligible papers with verified source passages, comparable measurements and temporal roles, and a direct match to the feature and prediction target. Each eligible paper \(\ell\) contributes at most one directional assessment per feature:
\begin{equation}
r_{mj\ell}=
\begin{cases}
+1, & \text{supports the model coefficient direction},\\
0, & \text{inconclusive},\\
-1, & \text{conflicts with the model coefficient direction}.
\end{cases}
\end{equation}
A paper receives \(r_{mj\ell}=0\) when its eligible findings are mixed, statistically nonsignificant, or directionally unclear. Papers without eligible comparable findings are excluded from \(N_{mj}\).

The coordinator combines directional consistency with a saturating measure of evidence sufficiency:

\begin{equation}
\begin{aligned}
&D_{mj} =
\begin{cases}
\dfrac{1}{2}
\left(
1+\dfrac{1}{N_{mj}}
\sum_{\ell=1}^{N_{mj}}r_{mj\ell}
\right),
& N_{mj}>0,\\[6pt]
0, & N_{mj}=0,
\end{cases}
\\[8pt]
&S_{mj} =
1-\exp\!\left(-N_{mj}/\lambda\right),
\qquad \lambda=2,
\\[8pt]
&e_{mj} = D_{mj}S_{mj}.
\end{aligned}
\end{equation}
Directional consistency equals 1 when all eligible papers support the coefficient direction, 0 when all conflict, and 0.5 when all are inconclusive. The sufficiency term discounts agreement based on a small number of papers. A feature without eligible evidence has \(D_{mj}=S_{mj}=e_{mj}=0\).

Feature-level support is aggregated using the magnitude of each feature's contribution to the patient's prediction:

\begin{equation}
\begin{aligned}
&E_m =
\frac{
\sum_{j\in\mathcal{K}_m}|a_{mj}|e_{mj}
}{
\sum_{j\in\mathcal{K}_m}|a_{mj}|
},
\\[8pt]
&U_{\mathrm{evidence}} =
1-\frac{1}{|\mathcal{M}|}
\sum_{m\in\mathcal{M}}E_m.
\end{aligned}
\end{equation}

Supporting papers increase a feature's evidence support, while conflicting papers reduce it. A feature without eligible papers receives zero support, but its weight remains in the calculation so that missing evidence is accounted for. If none of the selected features has eligible papers, or if all contribution weights are zero, we set \(E_m=0\). Overall, stronger literature support produces a higher \(E_m\) and lower evidence uncertainty, whereas missing or conflicting evidence produces lower support and higher uncertainty.

When all three components are available, the coordinator computes overall uncertainty and a complementary confidence index:
\begin{equation}
\begin{aligned}
U_{\mathrm{overall}} &=
\frac{
U_{\mathrm{pred}}
+
U_{\mathrm{conflict}}
+
U_{\mathrm{evidence}}
}{3},\\
C_{\mathrm{overall}} &= 1-U_{\mathrm{overall}}.
\end{aligned}
\end{equation}

All uncertainty indices range from 0 to 1, with higher values indicating greater uncertainty. A high overall uncertainty score corresponds to low confidence in the prediction, suggesting that the result should be interpreted cautiously and may warrant further assessment by a clinician. The coordinator reports each component separately so that clinicians can identify whether the uncertainty comes from a model score close to its classification threshold, disagreement among modality-specific agents, or insufficient or conflicting literature evidence.

\textbf{Human-in-the-Loop Review.}
The coordinator agent uses overall uncertainty to automatically identify predictions that warrant clinician review. When the overall uncertainty score is available, the review flag is defined as
\begin{equation}
H_{\mathrm{review}}
=
\mathbb{I}\!\left[
U_{\mathrm{overall}}>\tau_{\mathrm{review}}
\right],
\end{equation}
where \(\tau_{\mathrm{review}}\) is a predefined review threshold. A flag of \(H_{\mathrm{review}}=1\) prompts the coordinator to request clinician assessment of the intermediate findings underlying the prediction.

For cases with MRI data, the coordinator presents the ROIs flagged as abnormal, together with their anatomical labels, predicted abnormalities, and classifier scores. The clinician visually inspects these regions and can retain each original abnormality score or replace it with a reviewed binary value: 0 for normal and 1 for abnormal.

Once the clinician submits corrections, our system automatically updates the corresponding entries in the feature representation and trigger recomputation of affected predictions, literature assessments, uncertainty indicators, and reports. This human-in-the-loop process incorporates clinician corrections directly into the model inputs, allowing expert review to change the final progression score and potentially the predicted label. The agents automatically propagate these corrections, ensuring that clinician feedback is reflected in the updated assessment.

\textbf{Final Report Generation.}
The coordinator agent integrates the fusion-model prediction, modality-specific outputs, literature assessments, and uncertainty estimates into a structured multi-modal prediction report. The report presents the final outcome alongside its supporting model findings, external evidence, and unresolved discrepancies. It contains four components:
\begin{itemize}
    \item \textbf{Primary prediction:} The predicted outcome, Progressor or Non-progressor, together with the continuous LR progression score and the corresponding classification threshold.

    \item \textbf{Multimodal assessment:} Predictions from the available modality-specific agents and a summary of agreement or disagreement among them.

    \item \textbf{Model explanation:} The features contributing most strongly to the prediction, including their standardized values, logistic-regression coefficients, and contributions to the model log-odds.

    \item \textbf{Evidence and uncertainty assessment:} Literature support and conflicts concerning the influential feature--progression relationships, together with inconclusive findings, evidence gaps, and links to the source articles. The report also presents the overall uncertainty and its individual components to help clinicians identify findings that warrant further review.
\end{itemize}
The coordinator examines numerical results and evidence references against the structured tool outputs before presenting the report, maintaining consistency between the final assessment and the underlying agent outputs.

\section{Experiment}
\subsection{Dataset}

We use the Foundation for the National Institutes of Health (FNIH) Osteoarthritis Biomarkers Consortium cohort, a nested case--control sample of 600 participants from the Osteoarthritis Initiative, with one index knee per participant. The cohort includes 297 progressors and 303 non-progressors for each of two separately defined endpoints: Structural and pain progression. Structural progression is defined as a reduction of at least 0.7~mm in minimum medial tibiofemoral joint-space width from baseline to the 24-, 36-, or 48-month visit. Pain progression is defined as an increase of at least 9 points in the WOMAC pain score, normalized to a 0--100 scale, relative to baseline at the 24-, 36-, or 48-month visit. This increase must occur at two or more visits between months 24 and 60 to establish persistent worsening.

Baseline three-dimensional DESS MRI images, radiograph-derived measurements, and clinical and demographic variables are used to predict the two endpoints. A split stratified by the four FNIH progression groups reserves 100 knees for testing, with 50 progressors and 50 non-progressors for each endpoint. The remaining 500 knees are used to train progression prediction models. 


\subsection{Progression Prediction Performance}
\begin{table}[htbp]
    \centering
    \caption{AUROC for structural and pain progression prediction using different input modalities.}
    \label{tab:progression_auroc}
    \begin{tabular}{lcc}
        \toprule
        \textbf{Input} & \textbf{Structural AUROC} & \textbf{Pain AUROC} \\
        \midrule
        Clinical & 0.60 & 0.63 \\
        X-ray    & 0.62 & 0.58 \\
        MRI      & 0.80 & 0.58 \\
        Fusion   & \textbf{0.80} & \textbf{0.68} \\
        \bottomrule
    \end{tabular}
\end{table}

Table~\ref{tab:progression_auroc} reports the AUROC of the modality-specific and fusion models for structural and pain progression prediction. Fusion achieved the highest AUROC for both structural progression (0.80) and pain progression (0.68), in line with prior multi-modal KOA studies.

\subsection{Case Study: Multimodal Assessment of Structural Progression}
\label{sec:jsl-case-study}

\begingroup

\providecommand{\koapmid}[1]{%
  \href{https://pubmed.ncbi.nlm.nih.gov/#1/}{PMID~#1}}

\newcommand{\koaheading}[1]{%
  \par\medskip\noindent\textbf{#1}\par\nobreak\smallskip}

\setlength{\tabcolsep}{3pt}
\renewcommand{\arraystretch}{1.12}

\begin{tcolorbox}[
  enhanced,
  breakable,
  title={Structured Multimodal Prediction Report},
  colback=white,
  colframe=black!45,
  colbacktitle=black!7,
  coltitle=black,
  fonttitle=\bfseries,
  fontupper=\small,
  boxrule=0.5pt,
  arc=1mm,
  left=7pt,
  right=7pt,
  top=6pt,
  bottom=6pt,
  before skip=8pt,
  after skip=8pt
]

\textbf{Assessment:} Baseline\\
\textbf{Endpoint:} Structural progression


\koaheading{1. Primary prediction}

\textbf{Final classification: Progressor.}
The fusion logistic-regression (LR) score is
\textbf{0.8100}, above the training-derived threshold of
0.4505. 


\koaheading{2. Multimodal assessment}

\noindent
\begin{tabularx}{\linewidth}{@{}Xrrl@{}}
\toprule
Agent/model & LR score & Threshold & Prediction \\
\midrule
MRI agent
& 0.7580 & 0.4563 & Progressor \\

X-ray agent
& 0.5120 & 0.5320 & Non-progressor \\

Clinical agent
& 0.5524 & 0.5222 & Progressor \\

\textbf{Fusion model}
& \textbf{0.8100}
& \textbf{0.4505}
& \textbf{Progressor} \\
\bottomrule
\end{tabularx}

\smallskip
The MRI and clinical predictions agree with the final
classification. The X-ray score is 0.0200 below its threshold,
producing a discordant classification that is retained for
review. The final classification is determined by the fusion
model, rather than majority voting.


\koaheading{3. Model explanation}

Features are ranked by absolute contribution to model log-odds.
Here, $z_j$ expresses the patient's feature value relative to the training-set
mean in standard-deviation units, and $\beta_j$ is the learned
LR weight for that feature. Their product, $\beta_jz_j$, is the
feature's additive contribution to model log-odds: positive values
increase the prediction score, while negative values decrease it.

\koaheading{3.1. Fusion model: top five contributions}

{\footnotesize
\noindent
\begin{tabularx}{\linewidth}{
  @{}>{\raggedright\arraybackslash}Xrrr@{}
}
\toprule
Feature & $z_j$ & $\beta_j$ & $\beta_jz_j$ \\
\midrule
Kellgren--Lawrence grade
& $+1.170$ & $-0.492$ & $-0.576$ \\

Medial meniscus anterior horn abnormality probability
& $+3.408$ & $+0.147$ & $+0.501$ \\

Lateral femoral cartilage bone-scaled volume index
& $+1.793$ & $+0.194$ & $+0.348$ \\

Medial joint-space-narrowing grade
& $+1.158$ & $+0.248$ & $+0.288$ \\

Medial meniscal tibial-footprint coverage ratio
& $+0.695$ & $+0.350$ & $+0.244$ \\
\bottomrule
\end{tabularx}
\par}

\koaheading{3.2. MRI agent: top five contributions}

{\footnotesize
\noindent
\begin{tabularx}{\linewidth}{
  @{}>{\raggedright\arraybackslash}Xrrr@{}
}
\toprule
Feature & $z_j$ & $\beta_j$ & $\beta_jz_j$ \\
\midrule
Medial meniscus anterior horn abnormality probability
& $+3.408$ & $+0.075$ & $+0.255$ \\

Lateral femoral cartilage bone-scaled volume index
& $+1.793$ & $+0.099$ & $+0.178$ \\

Anterior medial tibial cartilage full-thickness-loss probability
& $+2.070$ & $+0.080$ & $+0.166$ \\

Medial meniscus body abnormality probability
& $+2.066$ & $+0.060$ & $+0.124$ \\

Anterior medial tibial cartilage lesion-presence probability
& $+1.494$ & $+0.077$ & $+0.115$ \\
\bottomrule
\end{tabularx}
\par}

\koaheading{3.3. X-ray agent: top five contributions}

{\footnotesize
\noindent
\begin{tabularx}{\linewidth}{
  @{}>{\raggedright\arraybackslash}Xrrr@{}
}
\toprule
Feature & $z_j$ & $\beta_j$ & $\beta_jz_j$ \\
\midrule
Kellgren--Lawrence grade
& $+1.176$ & $-0.537$ & $-0.631$ \\

Minimum medial joint-space width
& $-0.596$ & $+0.927$ & $-0.553$ \\

Medial joint-space-narrowing grade
& $+1.164$ & $+0.415$ & $+0.483$ \\

Femoral-tibial alignment angle
& $-3.164$ & $-0.131$ & $+0.415$ \\

Medial tibial sclerosis grade
& $+0.425$ & $+0.583$ & $+0.248$ \\
\bottomrule
\end{tabularx}
\par}

\koaheading{3.4. Clinical agent: top five contributions}

{\footnotesize
\noindent
\begin{tabularx}{\linewidth}{
  @{}>{\raggedright\arraybackslash}Xrrr@{}
}
\toprule
Feature & $z_j$ & $\beta_j$ & $\beta_jz_j$ \\
\midrule
Female sex indicator
& $-1.205$ & $-0.260$ & $+0.313$ \\

Baseline WOMAC physical function
& $+0.950$ & $+0.218$ & $+0.207$ \\

Baseline age
& $-0.992$ & $+0.209$ & $-0.207$ \\

Baseline body mass index
& $-1.792$ & $+0.070$ & $-0.125$ \\

Frequent knee-pain medication use
& $-0.652$ & $-0.045$ & $+0.029$ \\
\bottomrule
\end{tabularx}
\par}


\koaheading{4. Literature evidence assessment}
The agent checks whether PubMed findings are comparable with the model features and prediction task in terms of measurement, anatomy, timing, outcome. For comparable findings, it compares the reported association direction with the sign of the corresponding LR coefficient. Agreement indicates literature support for the model’s feature–progression relationship, but does not confirm whether an individual patient will progress.

{\footnotesize
\noindent
\begin{tabularx}{\linewidth}{
  @{}
  >{\raggedright\arraybackslash}p{0.13\linewidth}
  >{\raggedright\arraybackslash}X
  @{}
}
\toprule
\textbf{Modality} & \textbf{Evidence assessment} \\
\midrule

MRI
&
Support for medial meniscal abnormalities was qualified
by inconclusive findings and limited comparability.
The remaining MRI features lacked clear comparable support,
primarily due to differences in measurements or endpoints. \\

\addlinespace
X-ray
&
Evidence supported the model relationship for medial
joint-space-narrowing grade but conflicted with that for
minimum medial joint-space width. Findings for the
remaining features were inconclusive or not comparable. \\

\addlinespace
Clinical
&
BMI-related findings were inconclusive. Retrieved findings
for sex, age, physical function, and medication use were
classified as not comparable to the corresponding
model relationships. \\

\bottomrule
\end{tabularx}

\smallskip
\noindent
These assessments characterize literature support for
modality-specific feature relationships and inform
evidence uncertainty. They do not establish the correctness
of the patient-level fusion prediction.
\par}


\koaheading{5. Uncertainty assessment}

\noindent
\begin{tabularx}{\linewidth}{@{}Xr@{}}
\toprule
Component & Value \\
\midrule
Prediction uncertainty & 0.7814 \\
Modality-conflict uncertainty  & 0.0955 \\
Literature evidence uncertainty & 0.8596 \\
\midrule
\textbf{Overall uncertainty} & \textbf{0.5788} \\
Complementary confidence & 0.4212 \\
\bottomrule
\end{tabularx}




\koaheading{6. Human review and correction}

Findings warranting review include the X-ray cross modality disagreement,
the medial meniscal and anterior medial tibial cartilage
abnormalities, and feature relationships with conflicting or
insufficient evidence. Clinician corrections can be passed
through the feature-override workflow to update predictions
and the report. As the overall uncertainty exceeds the predefined threshold
of 0.5, the case is flagged as \textbf{Clinician Review Required}.

\end{tcolorbox}
\endgroup


\section{Discussion}
In this study, we present a multi-agent framework for knee osteoarthritis progression assessment, covering both structural and pain progression. The framework integrates MRI, radiographic, and clinical information through modality-specific agents and a coordinator agent. It automates patient-data retrieval, task-specific tool selection, progression prediction, literature evidence assessment, and structured report generation.

The framework connects automated prediction with interpretable findings, literature evidence, and clinician oversight. Feature-level evidence assessment identifies agreement, potential conflicts, and gaps between modelled relationships and literature findings. The coordinator reports these assessments alongside prediction uncertainty and cross-modal disagreement, providing a transparent basis for reviewing the final prediction that helps determine when a  prediction should be trusted, questioned, or interpreted cautiously. The human-in-the-loop workflow further allows clinicians to correct intermediate MRI findings, with the agents propagating these corrections through subsequent predictions and reports. This enables clinical feedback to directly influence the final assessment while retaining automated coordination of the analysis.

Future work will expand the framework by incorporating additional analysis tools and prediction models and extending its coverage to more clinical endpoints. These extensions will broaden the range of tasks supported by the agents within the similar coordinated workflow.

\bibliography{iclr2027_conference}
\bibliographystyle{iclr2027_conference}


\end{document}